# Self-Bounded Prediction Suffix Tree via Approximate String Matching

**Dongwoo Kim** [1 2]   **Christian Walder** [3 1]

## Abstract

Prediction suffix trees (PST) provide an effective tool for sequence modelling and prediction. Current prediction techniques for PSTs rely on exact matching between the suffix of the current sequence and the previously observed sequence. We present a provably correct algorithm for learning a PST with approximate suffix matching by relaxing the exact matching condition. We then present a self-bounded enhancement of our algorithm where the depth of suffix tree grows automatically in response to the model performance on a training sequence. Through experiments on synthetic datasets as well as three real-world datasets, we show that the approximate matching PST results in better predictive performance than the other variants of PST.

## 1. Introduction

Prediction suffix trees (PST) provide an elegant and effective tool for sequence prediction tasks such as compression, temporal classification, reinforcement learning, and DNA sequencing (Li & Fu, 2014; Majumdar, 2016; Messias & Whiteson, 2017). The advantage of PSTs over other fixed-order Markov model is that the number of symbols used to predict depends on prediction context through the suffix tree data structure, which provides an efficient way to store and retrieve a set of strings and all their suffixes (Bellemare et al., 2014).

Many PST algorithms perform *exact matching* between the suffix of the current sequence and sub-sequences in the previous sequence (Ron et al., 1996). The algorithms then make a prediction based on the previous history of those sub-sequences. Although the exact matching explicitly models the context where the same pattern occurred, it is potentially vulnerable to variation in a sequence such as substitution noise. To illustrate the prediction under variation, we present two symbolic music scores in Figure 1. When a PST model performs a prediction on the first score, it can take advantage of the same pattern between the first and last three symbols, where the latter is the suffix of the current sequence. On the other hand, the model cannot relate the first and last three symbols given the second score because of the minor variation in the suffix with the same length.

[1]Australian National University, Canberra, ACT, Australia [2]Data to Decisions CRC, Kent Town, SA, Australia [3]Data61 at CSIRO, Canberra, ACT, Australia. Correspondence to: Dongwoo Kim <dongwoo.kim@anu.edu.au>.

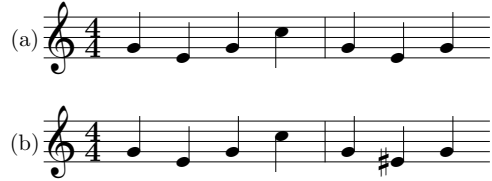

*Figure 1.* Musical note prediction: how do we predict the next symbol of each score? (a) A PST can relate the first and last three symbols and makes a prediction with the suffix of length three. (b) A PST cannot relate the first and last three symbols due to the little variation in the suffix with the same length although the pattern is similar.

Another practical assumption made in earlier work is that a shorter suffix has a higher priority than a longer one during prediction (Dekel et al., 2005; Karampatziakis & Kozen, 2009). This might be an appropriate assumption for some domains where the recently observed symbols are more important than the previous symbols, but this might be inappropriate for other domains. For example, a longer suffix would be more important than a suffix of length one to predict the next symbol of a music sheet since the temporal pattern in music is often continued over multiple notes.

In this paper, we provide a novel construction of the prediction suffix tree and its online learning algorithm via approximate string matching. In that sense, the proposed algorithm can be robust to small variations in a sequence. We also provide a mechanism to control the importance of different suffixes based on their length. Finally, we derive a self-bounded version of the proposed model that decides the maximum length of suffix based on a trade-off between a confidence of prediction and complexity of algorithm automatically.

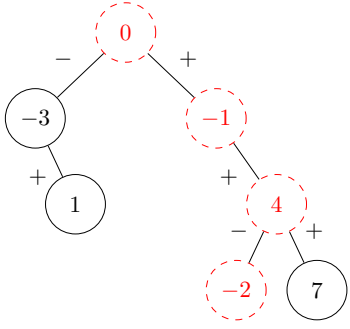

*Figure 2.* An example of binary prediction suffix tree $\mathcal{T} = \{-, +, +-, ++, -++, +++\}$ with score function $g$, i.e. $g(+-) = 1$. The dashed (red) nodes form suffix $-++$ and can be used to the next symbol of sequences having the same suffix such as $-++, --++, +-++$.

In the next section, we describe a decision theoretic PST model and how to learn the model parameters. In Section 3, we derive a PST with approximate string matching and its online learning algorithm. We also proof the mistake bound of the proposed algorithm w.r.t. an arbitrary hypothesis. In Section 4, we enhance our model by letting the model adaptively choose the depth of suffix tree. In Section 5 and 6, we verify our approach on synthetic datasets and demonstrate the improved predictive performance of our model on three real-world datasets.

## 2. Background

We start by providing a formal description of a decision theoretic PST model. Assume that an input stream is a sequence of vectors $\mathbf{x}_1, \mathbf{x}_2, ...(\mathbf{x}_t \in \mathbb{R}^n)$, and an output stream is a sequence of binary symbols $y_1, y_2, ...(y_t \in \{-1, +1\})$. We will relax the binary assumption in Subsection 5.2. We denote a sub-sequence of output $y_i, y_{i+1}, ..., y_j$ by $\mathbf{y}_i^j$. Our goal is to predict the next symbol $y_t$ given the binary sequence $\mathbf{y}_1^{t-1}$ and the next input vector $\mathbf{x}_t$

Dekel et al. propose a provably-correct PST algorithm to predict a sequence of symbols. With suffix-closed tree $\mathcal{T}$[1] endowed with a score at each node, the prediction function for symbol $y_t$ given $\mathbf{y}_1^{t-1}$ and $\mathbf{x}_t$ is

$$h(\mathbf{x}_t, \mathbf{y}_1^{t-1}) = \mathbf{w} \cdot \mathbf{x}_t + \sum_{i=1}^{t-1} 2^{-i/2} g(\mathbf{y}_{t-i}^{t-1}), \quad (1)$$

where $\mathbf{w} \in \mathbb{R}^n$ is a weight vector, and $g(\mathbf{s})$ is a score of node $\mathbf{s}$ in a suffix tree $\mathcal{T}$. The score of a suffix is zero if the tree does not have the suffix. We then use the sign of this prediction function as a predicted symbol $y_t$. The prediction function looks up scores of all possible suffixes of the input stream up to time $t-1$ from suffix tree $\mathcal{T}$ and takes a weighted sum of the scores of suffixes with *exponential decaying* weight $2^{-i/2}$ w.r.t. the length of the suffix. Finally, the weighted score is added to the inner product between weight vector $\mathbf{w}$ and input vector $\mathbf{x}$ to make a prediction. The magnitude of $h$ represents the confidence in this prediction.

[1]$\mathcal{T}$ is a suffix-closed if $\forall \mathbf{s} \in \mathcal{T}$, all suffixes of $\mathbf{s}$ are also in $\mathcal{T}$.

Figure 2 shows an example of prediction suffix tree with six suffixes therein. The value of node shows the score of a corresponding suffix, e.g., $g(-++) = 4$. Assume that we want to predict the next symbol of sequence $\mathbf{y}_1^4 = --++$, then, with the prediction function given Eq. 1, the predicted symbol of $y_5$ is $\text{sign}(2^{-1/2} \times (-1) + 2^{-2/2} \times (4) + 2^{-3/2} \times (-2))$.

There are multiple ways to construct the suffix tree and learn the model parameters. In this paper, we focus on the margin-based online learning algorithm and its analysis as per Dekel et al.; Karampatziakis & Kozen. In online learning, the model parameters are updated after each round. At round $t$, the model makes a prediction given input $\mathbf{x}_t$ and the previous sequence $\mathbf{y}_1^{t-1}$. The predicted symbol $\hat{y}_t = \text{sign}(h_t(\mathbf{x}_t, \mathbf{y}_1^{t-1}))$ is then compared to the revealed correct symbol $y_t$. Finally, the prediction function is updated based on the prediction and true symbol. More formally, the model computes the hinge loss after each round

$$\ell_t = \max\{0, 1 - y_t h_t(\mathbf{x}_t, \mathbf{y}_1^{t-1})\}. \quad (2)$$

Then the weight vector and node scores are updated based on the loss suffered from the prediction. The update rules for $\mathbf{w}$ and $g$ for all $\mathbf{s} \in \{\mathbf{y}_{t-i}^{t-1}\}_{i=1}^{t-1}$ are as follows:

$$\mathbf{w}_{t+1} = \mathbf{w}_t + y_t \tau_t \mathbf{x}_t \quad (3)$$

$$g_{t+1}(\mathbf{s}) = \begin{cases} g_t(\mathbf{s}) + y_t 2^{-|\mathbf{s}|/2} \tau_t, & \text{if } \mathbf{s} \in \mathcal{T}_t \\ y_t 2^{-|\mathbf{s}|/2} \tau_t, & \text{otherwise} \end{cases} \quad (4)$$

where $\tau_t$ depends on the loss $\ell_t$ and can be interpreted as a learning weight at time $t$, and $\mathcal{T}_t$ is a suffix tree at time $t$. When the updates happen, the unbounded version of this PST learning algorithm adds suffixes of the currently observed sequence into the suffix tree, resulting in the tree having $\mathcal{O}(t)$ depth and $\mathcal{O}(t^2)$ nodes. The same authors derive a self-bounded PST where the depth of tree is automatically chosen based on the model performance.

## 3. PST with Approximate String Matching

We propose a new prediction suffix tree algorithm with approximate string matching (aPST) where the prediction on next symbol depends on the exact matching as well as approximate matching suffixes. From the previous section, we observe that the PST algorithm has the following properties:

- The model performs exact matching between the current suffix and a node in the suffix tree where each node

represents sub-sequences of the previously observed sequence.

- The weight of shorter suffixes is higher than longer suffixes due to the exponential decaying rate.

As we have seen in Section 1, the first property prevents the model to take into account similar sub-sequences in the previous sequence, and the second property does not reflect the importance of suffix length. Our new prediction function takes into account all possible suffixes within $\epsilon$-Hamming distance from the original suffixes and provides a controllable weighting scheme. Formally, the prediction function of aPST is defined as:

$$h(\mathbf{x}_t, \mathbf{y}_1^{t-1}) = \mathbf{w}^\top \mathbf{x}_t + \sum_{i=1}^{t-1} \sum_{k=0}^{\epsilon} \sum_{\substack{\mathbf{s}: \mathbf{s} \in \mathcal{T}_{t-1}, \\ d(\mathbf{s}, \mathbf{y}_{t-i}^{t-1}) = k}} \omega(i,k) g(\mathbf{s}), \quad (5)$$

where $\omega : \mathbb{Z}^+ \times \mathbb{Z} \to \mathbb{R}^+$ controls the contribution of suffix $\mathbf{s}$, and $d(\mathbf{s}, \mathbf{s}')$ computes a distance between $\mathbf{s}$ and $\mathbf{s}'$ given some distance metric. We use Hamming distance throughout the paper, but the proposed framework may be applied to other distance metrics defined over strings. When $\epsilon = 0$, the model performs the exact string matching over the suffix tree. Hamming distance is only defined when the lengths of two sequences are the same (Robinson, 2003). Hence there is no suffix $\mathbf{s}$ whose distance $k$ from $\mathbf{y}_{t-i}^{t-1}$ is greater than $i$. We define the contribution function $\omega$ as

$$\omega(i,k) = \left( (1-\xi)^k \frac{\lambda^i \exp(-\lambda)}{i!} \right)^{\frac{1}{2}}, \quad \begin{array}{l} \lambda > 0 \\ 1 > \xi > 0 \end{array} \quad (6)$$

where $\lambda$ and $\xi$ are two model parameters. Unlike Eq. 1, where shorter sequences get a higher weight than longer ones, the model imposes different weights on different lengths of suffix. Specifically, the model gives the highest weight to suffixes of length $\lfloor \lambda \rfloor$. To reduce the contribution of approximate suffixes, we add an exponentially decaying factor w.r.t the distance from the original suffixes $k$.

The following theorem bounds the cumulative loss of the unbounded aPST online learning method of Algorithm 1 w.r.t. any fixed hypothesis $h^\star$ which could be chosen in hindsight. To derive the bound we define the norm of the score function $g$ as $||g||^2 = \sum_{\mathbf{s} \in \mathcal{T}} g(\mathbf{s})^2$.

**Theorem 1** (unbounded aPST). *Let $\mathbf{x}_1, \mathbf{x}_2, ..., \mathbf{x}_T$ be an input stream and let $y_1, y_2, ..., y_T$ be an output stream, where every $||\mathbf{x}||^2 \leq 1$ and every $y_t \in \{-1, +1\}$. Let $h^\star = (\mathbf{w}^\star, \mathcal{T}^\star, g^\star)$ be an arbitrary hypothesis such that $||g^\star||^2 < \infty$ and which attains the loss values $\ell_1^\star, \ell_2^\star, ... \ell_T^\star$. Let $\ell_1, ..., \ell_T$ be the sequence of losses attained by the unbounded online learning algorithm with the $\epsilon$-Hamming*

**Algorithm 1** Online learning algorithm for unbounded aPST.

1: **Input**: $\mathcal{T}_1 = \{\varnothing\}, \mathbf{w}_1 = \mathbf{0}, \lambda > 0, \epsilon \geq 0, 1 > \xi > 0$
2: Set $\gamma = \min\{-e^{-\lambda} + e^{\lambda(1-\xi)}, \frac{e^\lambda \Gamma(1+\epsilon,\lambda)}{\Gamma(1+\epsilon)}\}$
3: **for all** t = 1, 2, ..., T **do**
4:     Compute $h_t(\mathbf{x}_t, \mathbf{y}_1^{t-1})$
5:     Predict $\hat{y}_t = \text{sign}(h_t(\mathbf{x}_t, \mathbf{y}_1^{t-1}))$
6:     Receive $y_t$ and compute loss
        $\ell_t = \max\{0, 1 - y_t h_t(\mathbf{x}_t, \mathbf{y}_1^{t-1})\}$
7:     Set $\tau_t = \ell_t / (||\mathbf{x}||^2 + 2 + \gamma)$
8:     Set $d_t = t - 1$
9:     Update weight vector: $\mathbf{w}_{t+1} = \mathbf{w}_t + y_t \tau_t \mathbf{x}_t$
10:    Update tree:
11:       $\mathcal{T}_{t+1} = \mathcal{T}_t \cup \{\mathbf{y}_{t-i}^{t-1} : 1 \leq i \leq d_t\}$
12:       $g_{t+1}(\mathbf{s}) = g_t(\mathbf{s}) + y_t \tau_t \omega(|\mathbf{s}|, d(\mathbf{s}, \mathbf{y}_{t-i}^{t-1}))$
               if $\{\mathbf{s} : \mathbf{s} \in \mathcal{T}, d(\mathbf{s}, \mathbf{y}_{t-i}^{t-1}) \leq \epsilon\}$
13:       $g_{t+1}(\mathbf{s}) = y_t \tau_t \omega(|\mathbf{s}|, d(\mathbf{s}, \mathbf{y}_{t-i}^{t-1}))$
               if $\{\mathbf{s} : \mathbf{s} \notin \mathcal{T}, d(\mathbf{s}, \mathbf{y}_{t-i}^{t-1}) \leq \epsilon\}$
14: **end for**

*distance in Algorithm 1. Then it holds that*

$$\sum_{t=1}^T \ell_t^2 \leq \left( 3 + \gamma \right) \left( ||\mathbf{w}^\star||^2 + ||g^\star||^2 + \frac{1}{2} \sum_{t=1}^T (\ell_t^\star)^2 \right),$$

*where $\gamma = \min\{-e^{-\lambda} + e^{\lambda(1-\xi)}, \frac{e^\lambda \Gamma(1+\epsilon,\lambda)}{\Gamma(1+\epsilon)}\}$.*

*Proof.* Define $\Delta_t = ||\mathbf{w}_t - \mathbf{w}^\star||^2 - ||\mathbf{w}_{t+1} - \mathbf{w}^\star||^2$ and

$$\hat{\Delta}_t = \sum_{\mathbf{s} \in \mathcal{Y}^\star} (g_t(\mathbf{s}) - g^\star(\mathbf{s}))^2 - \sum_{\mathbf{s} \in \mathcal{Y}^\star} (g_{t+1}(\mathbf{s}) - g^\star(\mathbf{s}))^2. \quad (7)$$

We prove the theorem by using an upper and lower bound of $\sum_t (\Delta_t + \hat{\Delta}_t)$. First, expanding $\sum_t (\Delta_t + \hat{\Delta}_t)$ by the definition gives

$$\sum_t (\Delta_t + \hat{\Delta}_t) = ||\mathbf{w}_1 - \mathbf{w}^\star||^2 - ||\mathbf{w}_{t+1} - \mathbf{w}^\star||^2$$
$$+ \sum_{\mathbf{s} \in \mathcal{Y}^\star} \{(g_1(\mathbf{s}) - g^\star(\mathbf{s}))^2 - (g_{t+1}(\mathbf{s}) - g^\star(\mathbf{s}))^2\}. \quad (8)$$

Since $\mathbf{w}_1 = \mathbf{0}$ and $g_1(\cdot) = 0$, we can obtain a simple upper bound by omitting the negative terms

$$\sum_t (\Delta_t + \hat{\Delta}_t) \leq ||\mathbf{w}^\star||^2 + ||g^\star||^2, \quad (9)$$

where $||g^\star||^2 = \sum_{\mathbf{s} \in \mathcal{Y}^\star} g^\star(\mathbf{s})^2$. To derive the lower bound, we first rewrite $\Delta_t$ as $||\mathbf{w}_t - \mathbf{w}^\star||^2 - ||(\mathbf{w}_{t+1} - \mathbf{w}_t) + (\mathbf{w}_t - \mathbf{w}^\star)||^2$ by adding null term $\mathbf{w}_t - \mathbf{w}_t$. A further derivation gives $\Delta_t = -||\mathbf{w}_{t+1} - \mathbf{w}_t||^2 - 2(\mathbf{w}_{t+1} - \mathbf{w}_t) \cdot (\mathbf{w}_t - \mathbf{w}^\star)$. With the update rule $\mathbf{w}_{t+1} = \mathbf{w}_t + y_t \tau_t \mathbf{x}_t$, we can obtain

$$\Delta_t = -\tau_t^2 ||\mathbf{x}_t||^2 - 2y_t \tau_t \mathbf{x}_t (\mathbf{w}_t - \mathbf{w}^\star). \quad (10)$$

We manipulate the second term in Eq. 7 in a similar way by adding null term $g_t(\mathbf{s}) - g_t(\mathbf{s})$ to get

$$\hat{\Delta}_t = \sum_{\mathbf{s} \in \mathcal{Y}^\star} \left\{ \left(g_t(\mathbf{s}) - g^\star(\mathbf{s})\right)^2 \right.$$
$$\left. - \left((g_{t+1}(\mathbf{s}) - g_t(\mathbf{s})) + (g_t(\mathbf{s}) - g^\star(\mathbf{s}))\right)^2 \right\}$$
$$= \sum_{\mathbf{s} \in \mathcal{Y}^\star} \left\{ -\left(g_{t+1}(\mathbf{s}) - g_t(\mathbf{s})\right)^2 \right.$$
$$\left. - 2\left((g_{t+1}(\mathbf{s}) - g_t(\mathbf{s}))(g_t(\mathbf{s}) - g^\star(\mathbf{s}))\right) \right\}. \quad (11)$$

Note that the algorithm updates $g_t(\mathbf{s})$ at time $t$ only if $\mathbf{s}$ is one of the approximate suffixes of $\{\mathbf{y}_{t-i}^{t-1}\}_{i=1}^{d_t}$ with $d_t = t-1$. Therefore $g_{t+1}(\mathbf{s}) - g_t(\mathbf{s})$ would only have non-zero value if suffix $\mathbf{s}$ is within $\epsilon$-neighbourhood of $\mathbf{y}_{t-|\mathbf{s}|}^{t-1}$. Using this fact, $\hat{\Delta}_t$ can be further simplified as

$$\hat{\Delta}_t = \sum_{i=1}^{d_t} \sum_{k=0}^{\epsilon} \sum_{\substack{\mathbf{s}: \mathbf{s} \in \mathcal{T}_{t-1}, \\ d(\mathbf{s}, \mathbf{y}_{t-i}^{t-1}) = k}} -\tau^2 \omega(i, k)^2 \quad (12)$$
$$- 2 \sum_{i=1}^{d_t} \sum_{k=0}^{\epsilon} \sum_{\substack{\mathbf{s}: \mathbf{s} \in \mathcal{T}_{t-1}, \\ d(\mathbf{s}, \mathbf{y}_{t-i}^{t-1}) = k}} y_t \tau_t \omega(i, k)(g_t(\mathbf{s}) - g^\star(\mathbf{s})),$$

where we have used the update rule $g_{t+1}(\mathbf{s}) = g_t(\mathbf{s}) + y_t \tau_t \omega(i, k)$. By adding Eq. 10 and Eq. 12, we have that

$$\Delta_t + \hat{\Delta}_t = -\tau_t^2 \left( ||\mathbf{x}_t||^2 + \sum_{i=1}^{d_t} \sum_{k=0}^{\epsilon} \sum_{\substack{\mathbf{s}: \mathbf{s} \in \mathcal{T}_{t-1}, \\ d(\mathbf{s}, \mathbf{y}_{t-i}^{t-1}) = k}} \omega(i, k)^2 \right)$$
$$- 2\tau_t y_t \left( \mathbf{w}_t \mathbf{x}_t + \sum_{i=1}^{d_t} \sum_{k=0}^{\epsilon} \sum_{\substack{\mathbf{s}: \mathbf{s} \in \mathcal{T}_{t-1}, \\ d(\mathbf{s}, \mathbf{y}_{t-i}^{t-1}) = k}} \omega(i, k) g_t(\mathbf{s}) \right)$$
$$+ 2\tau_t y_t \left( \mathbf{w}^\star \mathbf{x}_t + \sum_{i=1}^{d_t} \sum_{k=0}^{\epsilon} \sum_{\substack{\mathbf{s}: \mathbf{s} \in \mathcal{T}_{t-1}, \\ d(\mathbf{s}, \mathbf{y}_{t-i}^{t-1}) = k}} \omega(i, k) g^\star(\mathbf{s}) \right). \quad (13)$$

Let $\gamma = \min\{-e^{-\lambda} + e^{\lambda(1-\xi)}, e^\lambda \Gamma(1+\epsilon, \lambda)/\Gamma(1+\epsilon)\}$. Combining Corollary 3.1 with the definition of $h_t$ and $h^\star$ leads to

$$\Delta_t + \hat{\Delta}_t \geq -\tau_t^2 \left( ||\mathbf{x}_t||^2 + \gamma \right)$$
$$- 2\tau_t y_t h_t(\mathbf{x}_t, \mathbf{y}_1^{t-1}) + 2\tau_t y_t h^\star(\mathbf{x}_t, \mathbf{y}_1^{t-1})$$
$$\geq -\tau_t^2 \left( ||\mathbf{x}_t||^2 + \gamma \right)$$
$$+ 2\tau_t(\ell_t - 1) + 2\tau_t(1 - \ell_t^\star) = \Psi_t. \quad (14)$$

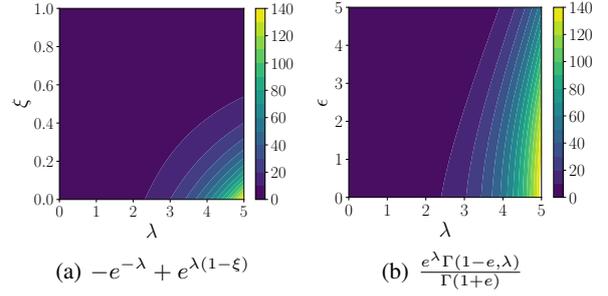

(a) $-e^{-\lambda} + e^{\lambda(1-\xi)}$  (b) $\frac{e^\lambda \Gamma(1-e,\lambda)}{\Gamma(1+e)}$

Figure 3. Contour plots of $\gamma$ used in Theorem 1.

If we subtract a non-negative term $(2^{1/2}\tau_t - 2^{-1/2}\ell_t^\star)^2$ from $\Psi_t$, then $\Psi_t$ is lower bounded by

$$\Psi_t \geq -\tau_t^2 \left( ||\mathbf{x}_t||^2 + 2 + \gamma \right) + 2\tau_t \ell_t - (\ell_t^\star)^2/2$$

Let $\tau_t = \ell_t/(||\mathbf{x}||^2 + 2 + \gamma)$. Using $||x||^2 \leq 1$,

$$\Psi_t \geq \tau_t \ell_t - \frac{(\ell_t^\star)^2}{2} = \frac{\ell_t^2}{||\mathbf{x}||^2 + 2 + \gamma} - \frac{(\ell_t^\star)^2}{2}$$
$$\geq \frac{\ell_t^2}{3 + \gamma} - \frac{(\ell_t^\star)^2}{2} \quad (15)$$

Combining the sum of lower bounds $\sum_{t=1}^{T}(\Delta_t + \hat{\Delta}_t)$ with the upper bound on $\sum_{t=1}^{T}(\Delta_t + \hat{\Delta}_t)$ in Eq. 9 gives us the bound stated in the theorem

$$||\mathbf{w}^\star||^2 + ||g^\star||^2 \geq \sum_{t=1}^{T} \left( \frac{\ell_t^2}{3+\gamma} - \frac{(\ell_t^\star)^2}{2} \right).$$

□

The competitive ratio bound obtained by Theorem 1 depends exponentially on $\lambda$ but also depends on the other parameter $\epsilon$ and $\xi$. We plot the contour maps of $\gamma$ within a sensible range of parameters in Figure 3. The contour suggests a plausible range of parameters to obtain competitive bound against an arbitrary hypothesis. For example, $\xi$ needs to be close to 1 if the algorithm gives more weight on long suffixes. It is also worth noting that if we set $\epsilon$ to 0, we can obtain the same competitive ratio provided in the original PST algorithm (Dekel et al., 2005) with more flexible Poisson weighting scheme.

The following lemmas used to prove Theorem 1 shows an upper bound on the squared sum of $\omega(i, k)$ given sequence $\mathbf{y}_1^{t-1}$ with respect to $\lambda$ and $\xi$.

**Lemma 2.** *Let $\omega(i, k) = ((1 - \xi)^k \lambda^i \exp(-\lambda)/i!)^{1/2}$. Given a binary sequence $\mathbf{y}_1^{t-1}$ and an arbitrary suffix tree $\mathcal{T}$ equipped with Hamming distance metric, $\sum_{i=1}^{t-1} \sum_{k=0}^{\epsilon} \sum_{\mathbf{s}: \mathbf{s} \in \mathcal{T}, d(\mathbf{s}, \mathbf{y}_{t-i}^{t-1}) = k} \omega^2(i, k) \leq -e^{-\lambda} +$*

$e^{\lambda(1-\xi)}\Gamma(t, -\lambda(-2+\xi))/\Gamma(t)$ *for all* $\lambda > 0$, $\epsilon \geq 0$, *and* $0 < \xi < 1$.

The proof of the lemma is provided in Appendix A.

The following simplification directly follows from the definition of the gamma function.

**Corollary 2.1.** *Under the assumption of Lemma 2,* $\sum_{i=1}^{t-1}\sum_{k=0}^{\epsilon}\sum_{\mathbf{s}:\mathbf{s}\in\mathcal{T}, d(\mathbf{s},\mathbf{y}_{t-i}^{t-1})=k} \omega^2(i,k) \leq -e^{-\lambda} + e^{\lambda(1-\xi)}$ *for all* $\lambda > 0$, $\epsilon \geq 0$, *and* $0 < \xi < 1$.

The bound in Lemma 2 depends on the value of $\lambda, t$ and $\xi$. We provide an alternative bound depending on $\lambda$ and $\epsilon$.

**Lemma 3.** *Let* $\omega(i,k) = ((1-\xi)^k \lambda^i \exp(-\lambda)/i!)^{1/2}$. *Given a binary sequence* $\mathbf{y}_1^{t-1}$ *and an arbitrary suffix tree* $\mathcal{T}$ *equipped with the Hamming distance metric,* $\sum_{i=1}^{t-1}\sum_{k=0}^{\epsilon}\sum_{\mathbf{s}:\mathbf{s}\in\mathcal{T}, d(\mathbf{s},\mathbf{y}_{t-i}^{t-1})=k} \omega^2(i,k) \leq e^{\lambda}\Gamma(1+\epsilon, \lambda)/\Gamma(1+\epsilon)$ *for all* $\lambda > 0$, $\epsilon \geq 0$, *and* $0 < \xi < 1$.

The proof of the lemma is also provided in Appendix B.

Finally, combining above lemmas provides the constant ratio used in Theorem 1.

**Corollary 3.1.** *Under the assumption of Corollary 2.1 and Lemma 3,* $\sum_{i=1}^{t-1}\sum_{k=0}^{\epsilon}\sum_{\mathbf{s}:\mathbf{s}\in\mathcal{T}, d(\mathbf{s},\mathbf{y}_{t-i}^{t-1})=k} \omega^2(i,k) \leq \min\{-e^{-\lambda} + e^{\lambda(1-\xi)}, \frac{e^{\lambda}\Gamma(1+\epsilon, \lambda)}{\Gamma(1+\epsilon)}\}$ *for all* $\lambda > 0$, $\epsilon \geq 0$, *and* $0 < \xi < 1$.

## 4. Self-Bounded aPST

The unbounded aPST algorithm relaxes the exact matching condition of PST. However, similarly to the unbounded PST, the depth of a suffix tree in the unbounded aPST also scales linearly in the length of sequence. In this section, we propose a self-bounded enhancement of the aPST algorithm which automatically grows a bounded-depth aPST. In each round, the algorithm decides whether to increase the depth of the suffix tree based on the prediction of next symbol.

The self-bounded aPST algorithm is described in Algorithm 2. We introduce tree depth variable $d_t$ calculated on every round of online iteration to represent the maximum depth of the suffix tree at time $t$. The proposed model automatically trades off between the size of suffix tree and confidence of prediction. We set the minimum value of $d_t$ to $\lceil \lambda + \epsilon \rceil$ in order to take into account the suffixes around the maximal weight length $\lfloor \lambda \rfloor$. Note that the algorithm updates the tree when the loss is greater than 1/2. This relaxed margin prevents the tree growing linearly with respect to the length of observed sequence. The following theorem provides the loss bound of proposed algorithm in exchange for having a relatively small aPST.

**Algorithm 2** Online learning algorithm for self-bounded aPST.
1: **Input**: $\mathcal{T} = \{\varnothing\}, \mathbf{w}_1 = \mathbf{0}, P_1 = 0, \lambda > 0, \epsilon \geq 0, \delta \in (0,1)$
2: Set $\bar{\Gamma} = \Gamma(1+\epsilon, \lambda)/\Gamma(1+\epsilon)$
3: Set $\gamma = \min\{-e^{-\lambda} + e^{\lambda(1-\xi)}, e^{\lambda}\bar{\Gamma}\}$
4: **for all** t = 1, 2, ..., T **do**
5:     Compute $h_t(\mathbf{x}_t, \mathbf{y}_1^{t-1})$
6:     Predict $\hat{y}_t = \text{sign}(h_t(\mathbf{x}_t, \mathbf{y}_1^{t-1}))$
7:     Receive $y_t$ and compute loss
       $\ell_t = \max\{0, 1 - y_t h_t(\mathbf{x}_t, \mathbf{y}_1^{t-1})\}$
8:     **if** $\ell > 1/2$ **then**
9:       Set $\tau_t = \ell_t/(||\mathbf{x}_t||^2 + 2 + \gamma)$
10:      Set $d_t = \max\{\lceil \lambda + \epsilon \rceil, d_{t-1}\}$
11:      Define function $f(d) = e^d \lambda^d d^{-d}$
12:      **while** $\bar{\Gamma} f(d_t) > \left(\frac{(P_{t-1}^2 + \tau_t \ell_t)^{1/2} - P_{t-1}}{2\tau_t}\right)^2$ **do**
13:        $d_t = d_t + 1$
14:      **end while**
15:      Set $P_t = P_{t-1} + 2\tau_t \sqrt{\bar{\Gamma} f(d_t)}$
16:      Update weight vector: $\mathbf{w}_{t+1} = \mathbf{w}_t + y_t \tau_t \mathbf{x}_t$
17:      Update suffix tree:
18:        $g_{t+1}(\mathbf{s}) = g_t(\mathbf{s}) + y_t \tau_t \omega(|\mathbf{s}|, d(\mathbf{s}, \mathbf{y}_{t-i}^{t-1}))$
         if $\{\mathbf{s} : \mathbf{s} \in \mathcal{T}, d(\mathbf{s}, \mathbf{y}_{t-i}^{t-1}) \leq \epsilon\}$
19:        $g_{t+1}(\mathbf{s}) = y_t \tau_t \omega(|\mathbf{s}|, d(\mathbf{s}, \mathbf{y}_{t-i}^{t-1}))$
         if $\{\mathbf{s} : \mathbf{s} \notin \mathcal{T}, d(\mathbf{s}, \mathbf{y}_{t-i}^{t-1}) \leq \epsilon\}$
20:     **else**
21:      Set: $\tau_t = 0, P_t = P_{t-1}$
22:     **end if**
23: **end for**

**Theorem 4** (Self-bounded aPST). *Let* $\mathbf{x}_1, \mathbf{x}_2, ..., \mathbf{x}_T$ *be an input stream and let* $y_1, y_2, ..., y_T$ *be an output stream, where every* $||\mathbf{x}||^2 \leq 1$ *and every* $y_t \in \{-1, +1\}$. *Let* $h^\star = (\mathbf{w}^\star, \mathcal{T}^\star, g^\star)$ *be an arbitrary hypothesis such that* $||g^\star||^2 \leq \infty$ *and which attains the loss values* $\ell_1^\star, \ell_2^\star, ... \ell_T^\star$. *Let* $\ell_1, ..., \ell_T$ *be the sequence of losses attained by the self-bounded online learning algorithm with the $\epsilon$-Hamming distance in Algorithm 2. Then the sum of square losses on those rounds where* $\ell_t > \frac{1}{2}$ *is bounded by*

$$\sum_{t:\ell_t > \frac{1}{2}} \ell_t^2 \leq \bar{\lambda}\left(\frac{1+\sqrt{5}}{2}||g^\star|| + ||\mathbf{w}^\star|| + \left(\frac{1}{2}\sum_{t=1}^T (\ell_t^\star)^2\right)^{\frac{1}{2}}\right)^2,$$

*where* $\bar{\lambda} = 3 + \min\{-e^{-\lambda} + e^{\lambda(1-\xi)}, \frac{e^{\lambda}\Gamma(1+\epsilon, \lambda)}{\Gamma(1+\epsilon)}\}$.

See Appendix C for the detailed proof. Here we provide a sketch of proof. Again, the proof starts from the upper bound and lower bound on $\Delta + \tilde{\Delta}$ defined in Theorem 1. The upper bound remains the same. The derivation of the lower bound is the same up to Eq. 13, however, by the definition of $h$, we need to add null terms consisting scores of the suffixes from $d_t + 1$ to $t - 1$ in $\mathcal{T}^\star$ to formulate the lower bound as a linear function of $h_t$ and $h^\star$ as done in Eq. 14. Adding null terms results in a remainder after reformulation. The lower bound on this remainder can be obtained by Chernoff bound since the weights of additional term can be bounded by the sum of Poisson tail distribution. The combination of two lower bounds leads to a new lower bound on $\Delta + \hat{\Delta}$. Given the combination of upper and lower bounds on $\Delta + \hat{\Delta}$, we further show that if $d_t$ satisfies the condition described in Algorithm 2, the sum of squared losses has the lower bound explained in Theorem 4 via mathematical induction.

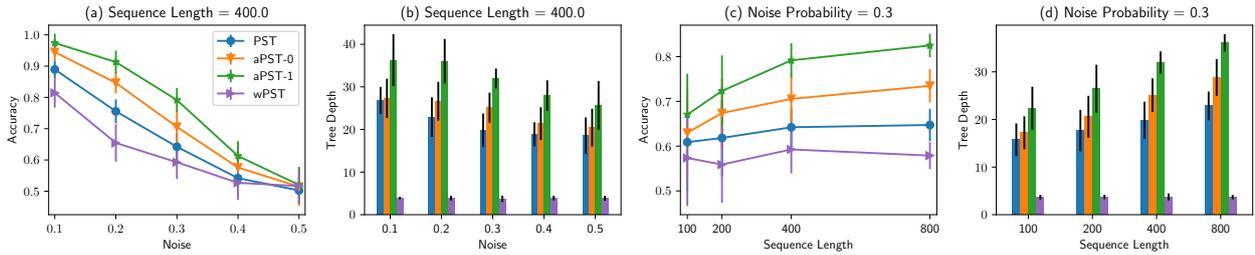

*Figure 4.* (a), (b): Prediction accuracy and tree depth of PSTs with respect to varying proportions of noise given the sequence length of 400. Error bar represents one standard deviation. As the proportion of noise increases, the prediction accuracy of both models decrease, and the accuracy converges to 0.5 when a half of sequence is corrupted by a random noise. Except that unpredictable case, aPST always outperforms PST while keeping a slightly larger suffix tree. The similar results are also observed with the other sequence lengths. (c), (d): Prediction accuracy and tree depth of PSTs with respect to varying lengths of sequence given the fixed noise level. As the sequence length increases, the accuracy of both models also increase in general.

## 5. Simulation Study

In this section, we compare the performance of aPST on a sequence prediction task to the classical PST (Dekel et al., 2005) and its variant wPST (Karampatziakis & Kozen, 2009) on a synthetic dataset. We start from a sequence motif, which is frequently occurred subsequences of an original sequence. Many sequences observed in real world applications can be rendered by a small number of motifs (Bailey et al., 2006; Ross et al., 2012). It is known that the PSTs are capable of identifying those motifs from a sequence (Majumdar, 2016). Given a sequence motif, we generate a random sequence with some level of noise, and then, compute predictive accuracies of PST algorithms to compare. Throughout this experiments, we focus on a situation where the input stream is unavailable, i.e. $\mathbf{x}_t = 0$ for all $t$ in a sequence, since the modelling on the input stream of aPST algorithm remains the same as those of PST. In other words, we measure and compare a sequence memorisation perspective of both models under a noisy environment in order to emphasise the importance of the approximate matching and weighting scheme.

### 5.1. Binary sequence prediction

We first synthesise simple sequence from the motif $[-1, -1, +1, +1]$. We construct a sequence by repeating the motif multiple times (25, 50, 100, 200 times), and then, we randomly corrupt each entry of input $y_i$ via an independent Bernoulli trial with a fixed noise probability. Without noise in a final sequence, all three models can predict perfectly after observing a first few entries. When a sequence is corrupted by some random noise, PST only relies on an input $\mathbf{x}_t$ if available, while aPST retrieves approximate suffixes to predict the next symbol.

For every experiment, we use the first 40% of a sequence to train, the subsequent 20% of the sequence to validate, and the final 40% of sequence to test the models. For both parameter $\lambda$ and $\epsilon$, we test all possible configuration of $\lambda = \{2, 4, 6, 8, 10, 12\}$, $\xi = (0.5, 0.7, 0.9, 0.99)$, and $\epsilon = \{0, 1\}$ and choose the best model based on the accuracy of validation set. All the experiments are repeated over 20 times with randomly corrupted entries. We report two quantities: accuracy of prediction on the uncorrupted entries and the final depth of the suffix tree.

Figure 4 (a) and (b) show the prediction accuracy and tree depth of the three different models with respect to varying proportions of noise. In general, as the proportion of noise increases, the prediction accuracy of both models decrease. The accuracy converges to the random baseline when the sequence is unpredictable, i.e. a half of sequence is corrupted. Aside from this unpredictable case, aPST always outperforms PST while keeping similar tree depths. Although wPST maintains the shallowest tree depth, it shows the lowest performance among all models.

Figure 4 (c) and (d) show the prediction accuracy and tree depth of the three different models with respect to varying lengths of sequence. The model predicts better when the sequence length is longer in general. The tree depth of aPST increases sub-linearly as the length of sequence increases. We can also observe that the variance of accuracy decreases as the sequence length increases with aPST.

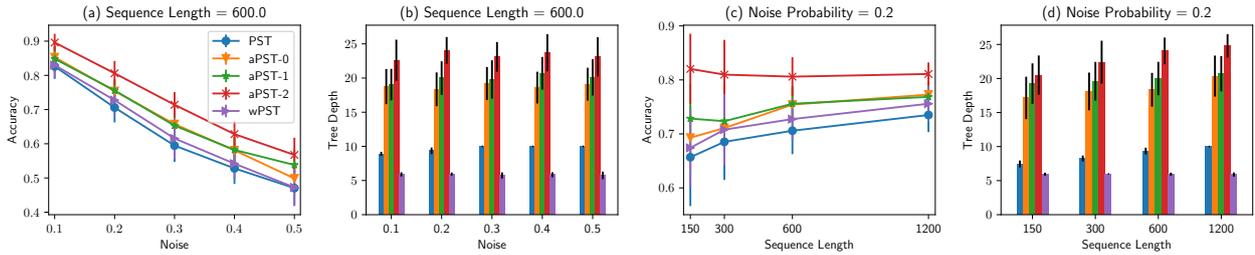

*Figure 5.* (a), (b): Prediction accuracy and tree depth of multiclass PSTs with respect to varying proportions of noise given the sequence length of 600. (c), (d): Prediction accuracy and tree depth of multiclass PSTs with respect to varying lengths of sequence given the fixed noise level. In general, the aPST outperforms the other models, and performs better when we increase the maximum Hamming distance $\epsilon$ between two sequences.

We present a more comprehensive analysis of the binary sequence prediction and demonstrate the performance on more complex synthetic sequences in Appendix D.

### 5.2. Multi-class sequence prediction

To predict on sequences with multiple symbols, we adopt ideas from (Crammer & Singer, 2001) and maintain trees $\mathcal{T}^{(1)}, \ldots \mathcal{T}^{(k)}$ for each symbol. The decision at time $t$ is $\hat{y}_t = \arg\max_k h_t^{(k)}(\mathbf{x}_t, \mathbf{y}_1^{t-1})$. If the prediction is wrong, we update the tree parameters of predicted symbol and true symbol with a piecewise margin loss defined as $\ell_t = \max_k \{h_t^k + 1 - h_t^{y_t}\}$ if $\hat{y}_t \neq y_t$. Hence, different trees might have different depth. Here, we report the maximum depth among the trees.

Note that there might be a combinatorial number of approximate suffixes if we add all approximate suffixes while updating the tree. To reduce the computational burden, we add the suffixes of the current sequence into the tree if the suffixes are not in the tree and update the approximate suffixes which are already in the suffix tree. Therefore, the suffix tree only contains the sub-sequences which have been observed in the past. The prediction still requires to search the approximate sequences over the suffix tree, but it can be done in an efficient way (Ukkonen, 1993; Giegerich & Kurtz, 1997).

For the experiment, we generate a random sequence from motif $[1, 2, 3, 4, 1, 3]$. Again, we inject random noise based on a Bernoulli trial with a fixed probability. The corrupted symbols are then replaced by random symbol with uniform probability over symbols. We follow the same experimental procedure as used in the binary experiments with the same set of parameters except that we tested $\epsilon$ up to 2.

Figure 5 shows the result of the synthetic experiments. In general, aPST outperforms both PST and wPST, and the accuracy increases as $\epsilon$ increases from 0 to 2. Again, this results emphasis the importance of approximate matching of PST under some noise in a sequence.

## 6. Experiments

In this section, we conduct some experiments with real datasets to demonstrate the practical effectiveness of the proposed method. We use three datasets: a symbolic music dataset (Walder, 2016) from which we retain midi onset events only, a system call dataset (Hofmeyr et al., 1998), and human activity dataset (Ordónez et al., 2013). The symbolic music dataset contains four sets of midi music dataset from different sources. The models predict a sequence of midi note number, which ranges 0-127. The system call dataset records a set of system call traces made by active processes, which might contain some intrusions of malicious programs. The models predict the next system call given a previous call sequence. The human activity dataset records a sequence of activities from two subjects. The models predict the next activity of each subject given a trajectory of activities.

We again compare aPST with PST and wPST in terms of prediction accuracy and tree depth. For every experiment, we use the first 30% of symbols to adjust the model parameters and use remaining 70% to report the model accuracy and tree depth. We use the same set of parameters used in the previous section.

Table 1, 2, 3 show the accuracies and final tree depth of PST models on music, system call, human activity datasets, respectively. For the music and human activity datasets, aPST outperforms the other models in terms of accuracy while using a slightly larger suffix tree. For the system call dataset, the tree depth of aPST is shallower than those of the other models while having a similar or better accuracies. The performance gain of aPST against the other models are

|     | PST  |       | wPST |       | aPST  |       |
| --- | ---- | ----- | ---- | ----- | ----- | ----- |
|     | Acc. | $d_T$ | Acc. | $d_T$ | Acc.  | $d_T$ |
| JBM | 0.203 | 6.2 | 0.231 | 6.0 | **0.246** | 8.5 |
| PMD | 0.349 | 7.5 | 0.386 | 5.9 | **0.442** | 10.4 |
| NOT | 0.469 | 6.7 | 0.510 | 5.9 | **0.571** | 9.9 |
| MUS | 0.237 | 6.6 | 0.253 | 5.9 | **0.270** | 8.7 |

Table 1. Average accuracy and tree depth of musical note prediction on four different music sources. aPST outperforms the other models in terms of accuracy with larger suffix trees.

|     | PST  |       | wPST |       | aPST  |       |
| --- | ---- | ----- | ---- | ----- | ----- | ----- |
|     | Acc. | $d_T$ | Acc. | $d_T$ | Acc.  | $d_T$ |
| [100, 200] | 0.292 | 7.0 | 0.287 | 6.0 | **0.300** | 5.1 |
| [200, 300] | 0.408 | 7.0 | 0.399 | 6.0 | **0.411** | 5.0 |
| [300, 500] | **0.803** | 7.0 | 0.801 | 6.0 | **0.803** | 5.2 |

Table 2. Average accuracy and tree depth of system call prediction broken down by the length of program sequence. The tree depths of aPST are shallower than those of the others while keeping similar or better accuracies.

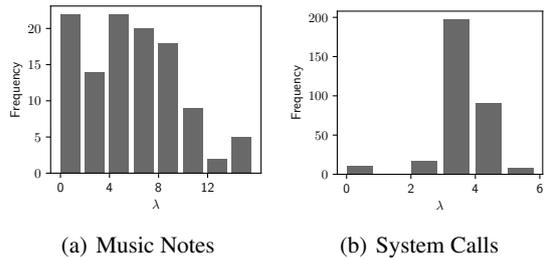

(a) Music Notes    (b) System Calls

Figure 6. Best $\lambda$ values obtained from two different datasets. A proper value of $\lambda$ varies across domains as well as sequences.

|     | PST  |       | wPST |       | aPST  |       |
| --- | ---- | ----- | ---- | ----- | ----- | ----- |
|     | Acc. | $d_T$ | Acc. | $d_T$ | Acc.  | $d_T$ |
| Subject A | 0.362 | 9 | 0.333 | 4 | **0.385** | 10 |
| Subject B | 0.191 | 9 | **0.238** | 4 | **0.238** | 10 |

Table 3. Accuracy and tree depth of human activity prediction.

significant when the sequence lengths are relatively short.

We plot the histogram of the best $\lambda$ values for the music dataset and system call dataset in Figure 6. The graph shows a quite distinctive characteristic of these two datasets. $\lambda$ values from the musical sequences distribute evenly across range 0 to 8, while $\lambda$ from system call traces are highly focused on the range $[3, 5)$. We conjecture that the songs have a greater variety of relevant motif lengths than the system call traces which have more static transition patterns (Nikolopoulos & Polenakis, 2014), therefore $\lambda$ values are adjusted according to the nature of a song.

## 7. Related Work

Variants of the PST algorithm have been developed in different scientific communities in different forms such as the variable length Markov models and context tree weighting (Willems et al., 1995; Helmbold & Schapire, 1997; Bühlmann et al., 1999; Bellemare et al., 2014). Most of these algorithms need an a-priori bound on the maximum number of previous symbols. Apostolico & Bejerano show that the upper bound assumption can be relaxed by a linear time prediction tree construction algorithm where the depth of suffix tree can increase up to the length of a training sequence. Dekel et al. propose an alternative self-bounded PST learning algorithm where the depth of prediction tree is bounded automatically based on a number of mistakes made by the algorithm. Karampatziakis & Kozen combine the idea of self-bounded PST and Winnow algorithm and derive a multiplicative update algorithm for online learning. By using the multiplicative update rules, the suggested algorithm can quickly adopt variation in complex sequence which exhibit different patterns at various points during life time. Xiao & Eckert further extend the PST to incorporate additional side information. They derive a second order online learning algorithm to take into account the variance of the estimator.

## 8. Conclusion

We have presented the decision theoretic prediction suffix tree with approximate string matching (aPST) by relaxing the exact matching condition of the PST models. The depth of the suffix tree generated by the proposed algorithm scales linearly with the length of the input sequence. To limit the depth of aPST, we proposed self-bounded version of aPST which automatically determines the depth of suffix tree. The loss bounds for both unbounded- and bounded-aPST are analysed. We showed that the applications of aPST to sequence modelling outperform the other PST models under a noisy environment via synthetic datasets. Furthermore, we showed the improved predictive performance of aPST on three real world datasets. Future work on this research will explorer wide range of distance metrics instead of the Hamming distance in order to take into account more complex editing behaviour in sequences.

## Acknowledgements

We thank Minjeong Shin, Cheng Soon Ong and Lexing Xie for instructive discussions on an early draft of this work. We also thank the anonymous reviewers for their detailed and thoughtful comments.

# Appendix for Self-Bounded Prediction Suffix Tree via Approximate String Matching

## A. Proof of Lemma 2

**Lemma 5.** *Let $\omega(i,k) = ((1-\xi)^k \lambda^i \exp(-\lambda)/i!)^{1/2}$. Given binary sequence $\mathbf{y}_1^{t-1}$ and arbitrary suffix tree $\mathcal{T}$ equipped with Hamming distance metric, $\sum_{i=1}^{t-1} \sum_{k=0}^{\epsilon} \sum_{\mathbf{s}:\mathbf{s}\in\mathcal{T}, d(\mathbf{s},\mathbf{y}_{t-i}^{t-1})=k} \omega^2(i,k) \leq -e^{-\lambda} + e^{\lambda(1-\xi)}\Gamma(t, -\lambda(-2+\xi))/\Gamma(t)$ for all $\lambda > 0$, $\epsilon \geq 0$, and $0 < \xi < 1$.*

*Proof.* Given a binary sequence of length $i$, there are at most $\binom{i}{k}$ possible suffixes that are exactly $k$-Hamming distance away from the original sequence if $i \geq k$, otherwise 0.[2] Therefore the sum of $\omega$ can be bounded by the number of possible approximate suffixes as

$$\sum_{i=1}^{t-1} \sum_{k=0}^{\epsilon} \sum_{\substack{\mathbf{s}:\mathbf{s}\in\mathcal{T},\\ d(\mathbf{s},\mathbf{y}_{t-i}^{t-1})=k}} \omega^2(i,k) \leq \sum_{i=1}^{t-1} \sum_{k=0}^{\epsilon} \binom{i}{k} \omega^2(i,k)$$

$$\leq \sum_{i=1}^{t-1} \frac{e^{-\lambda}(\lambda(2-\xi))^i}{i!}$$

$$= -e^{-\lambda} + \frac{e^{\lambda(1-\xi)}\Gamma(t, -\lambda(-2+\xi))}{\Gamma(t)} \quad (16)$$

where $\Gamma(a)$ and $\Gamma(a,b)$ are Gamma function and incomplete Gamma function, respectively. □

## B. Proof of Lemma 3

**Lemma 6.** *Let $\omega(i,k) = ((1-\xi)^k \lambda^i \exp(-\lambda)/i!)^{1/2}$. Given binary sequence $\mathbf{y}_1^{t-1}$ and arbitrary suffix tree $\mathcal{T}$ equipped with Hamming distance metric, $\sum_{i=1}^{t-1} \sum_{k=0}^{\epsilon} \sum_{\mathbf{s}:\mathbf{s}\in\mathcal{T}, d(\mathbf{s},\mathbf{y}_{t-i}^{t-1})=k} \omega^2(i,k) \leq e^{\lambda}\Gamma(1+\epsilon, \lambda)/\Gamma(1+\epsilon)$ for all $\lambda > 0$, $\epsilon \geq 0$, and $0 < \xi < 1$.*

*Proof.* Again, from the proof of Lemma 2, we can show

$$\sum_{i=1}^{t-1} \sum_{k=0}^{\epsilon} \binom{i}{k} \omega^2(i,k) = \sum_{i=1}^{t-1} \sum_{k=0}^{\epsilon} \frac{\lambda^k}{k!} \frac{\lambda^{i-k}\exp(-\lambda)}{(i-k)!} \mathbf{1}[i \geq k]$$

$$= \sum_{k=0}^{\epsilon} \sum_{i=1}^{t-1} \frac{\lambda^k}{k!} \frac{\lambda^{i-k}\exp(-\lambda)}{(i-k)!} \mathbf{1}[i \geq k]$$

$$\leq \sum_{k=0}^{\epsilon} \frac{\lambda^k}{k!}$$

$$= \frac{e^{\lambda}\Gamma(1+\epsilon, \lambda)}{\Gamma(1+\epsilon)} \quad (17)$$

where we use $\sum_{i=0}^{\infty} \lambda^i \exp(-\lambda)/i! = 1$. □

## C. Proof of Theorem 4

*Proof.* We use the same definition of $\Delta_t$ and $\hat{\Delta}_t$ as Theorem 1. The upper bound on $\sum_{t=1}(\Delta_t + \hat{\Delta}_t)$ in Eq. 9 and the equality on $\Delta_t + \hat{\Delta}_t$ in Eq. 13 still hold.

---
[2] Let $\binom{i}{k} = 0$ if $k \geq i$.

Let $\gamma = \min\{-e^{-\lambda} + e^{\lambda - \lambda\xi}, e^{\lambda}\Gamma(1+\epsilon, \lambda)/\Gamma(1+\epsilon)\}$. Given Eq. 13, Corollary 3.1 with the definition of $h_t$ and $h^\star$ gives

$$\begin{aligned}\Delta_t + \hat{\Delta}_t \geq & -\tau_t^2\Big(||\mathbf{x}_t||^2 + \gamma\Big) \\ & - 2\tau_t y_t h_t(\mathbf{x}_t, \mathbf{y}_1^{t-1}) + 2\tau_t y_t h^\star(\mathbf{x}_t, \mathbf{y}_1^{t-1}) \\ & - 2\tau_t y_t \sum_{k=0}^{\epsilon}\sum_{i=d_t+1}^{t-1}\sum_{\substack{\mathbf{s}: \mathbf{s}\in\mathcal{T}^\star, \\ d(\mathbf{s}, \mathbf{y}_{t-i}^{t-1})=k}} \omega(i,k) g^\star(\mathbf{s}),\end{aligned} \quad (18)$$

where we subtract the last term after constructing $h^\star$ by its definition in Eq. 5. The magnitude of the last summations can be bounded by the Cauchy-Schwartz inequality

$$\begin{aligned}\Big|\sum_{k=0}^{\epsilon}\sum_{i=d_t+1}^{t-1}&\sum_{\substack{\mathbf{s}: \mathbf{s}\in\mathcal{T}^\star, \\ d(\mathbf{s}, \mathbf{y}_{t-i}^{t-1})=k}}\omega(i,k)g^\star(\mathbf{s})\Big| \\ &\leq \Big(\sum_{k=0}^{\epsilon}\sum_{i=d_t+1}^{t-1}\binom{i}{k}\omega(i,k)^2\Big)^{1/2}||g^\star|| \\ &= \Big(\sum_{k=0}^{\epsilon}\sum_{i=d_t+1}^{t-1}\frac{\lambda^k}{k!}\frac{\lambda^{i-k}\exp(-\lambda)}{(i-k)!}\Big)^{1/2}||g^\star||.\end{aligned} \quad (19)$$

Given that $d_t \geq \lceil \lambda + \epsilon \rceil$, the Chernoff bound (Hoeffding, 1963) gives an upper bound on the square root

$$\begin{aligned}\sum_{k=0}^{\epsilon}\sum_{i=d_t+1}^{t-1}\frac{\lambda^k}{k!}\frac{\lambda^{i-k}\exp(-\lambda)}{(i-k)!} &\leq \sum_{k=0}^{\epsilon}\frac{\lambda^k}{k!}e^{d_t - \lambda}\lambda^{d_t}d_t^{-d_t} \\ &= \frac{\Gamma(1+\epsilon, \lambda)}{\Gamma(1+\epsilon)}e^{d_t}\lambda^{d_t}d_t^{-d_t}\end{aligned}$$

Alternatively, we may use a tighter bound of the Poisson tail distribution (Glynn, 1987). For now, let $\bar{\Gamma} = \Gamma(1+\epsilon, \lambda)/\Gamma(1+\epsilon)$ and $u_\lambda(d_t) = \bar{\Gamma}e^{d_t}\lambda^{d_t}d_t^{-d_t}$. Plugging the upper bound into Eq. 18 and combining the lower bound of $\Psi_t$ in Theorem 1 lead to

$$\begin{aligned}\Delta_t + \hat{\Delta}_t &\geq \Psi_t - 2\tau_t\sqrt{u_\lambda(d_t)}||g^\star|| \\ &\geq \tau_t \ell_t - \frac{1}{2}(\ell^\star)^2 - 2\tau_t\sqrt{u_\lambda(d_t)}||g^\star||\end{aligned} \quad (20)$$

Summing the lower bound over $t$ and comparing to the upper bound in Eq. 9 yield

$$L_T \leq ||g^\star||^2 + ||w^\star||^2 + \frac{1}{2}\sum_{t=1}^{T}(\ell_t^\star)^2 + ||g^\star||P_t \quad (21)$$

where $P_t = \sum_{i=1}^{t} 2\tau_t \sqrt{u_\lambda(d_t)}$ and $L_t = \sum_{i=1}^{t} \tau_t \ell_t$.

We now use mathematical induction to prove that $P_t^2 \leq L_t$. Assume $P_{t-1}^2 \leq L_{t-1}$, and let $P_0 = L_0 = 0$. By the definition of $P_t$, we can expand

$$\begin{aligned}P_t^2 &= (P_{t-1} + 2\tau_t\sqrt{u_\lambda(d_t)})^2 \\ &= P_{t-1}^2 + 4\tau_t\sqrt{u_\lambda(d_t)}P_{t-1} + 4\tau_t^2 u_\lambda(d_t)\end{aligned} \quad (22)$$

If we choose minimum $d_t$ which satisfies both $u_\lambda(d_t) \leq \big(((P_{t-1}^2 + \tau_t\ell_t)^{1/2} - P_{t-1})/2\tau_t\big)^2$ and $d_t \geq \lceil \lambda + \epsilon \rceil$, and plug this into Eq. 22, then with the inductive assumption we can show

$$P_t^2 \leq P_{t-1}^2 + \tau_t \ell_t \leq L_{t-1} + \tau_t \ell_t = L_t, \quad (23)$$

which proves the inductive argument. Note that the upper bound $u_\lambda(d_t)$ is strictly decreasing when $d_t \geq \lambda$, so we can always find the minimum $d_t$ which satisfies both conditions when $\ell > 1/2$. Since $P_t$ and $L_t$ are always positive, we have $P_T \leq \sqrt{L_T}$. Combining this inequality with Eq. 21 leads to

$$\left(\sqrt{L_T}\right)^2 \leq \|g^\star\|\sqrt{L_T} + \|g^\star\|^2 + \|\mathbf{w}^\star\|^2 + \frac{1}{2}\sum_{i=1}^T (\ell_t^\star)^2.$$

This equation is a quadratic inequality in $\sqrt{L_T}$. From the positive root of the quadratic equation, we get that

$$\sqrt{L_T} \leq \frac{1}{2}\left(\|g^\star\| + \left(5\|g^\star\|^2 + 4\|\mathbf{w}^\star\|^2 + 2\sum_{t=1}^T (\ell_t^\star)^2\right)^{\frac{1}{2}}\right).$$

Since $\sqrt{a^2 + b^2} \leq (a+b), (a,b \geq 0)$, the upper bound on $\sqrt{L_T}$ can be rewritten as

$$\sqrt{L_T} \leq \frac{1+\sqrt{5}}{2}\|g^\star\| + \|\mathbf{w}^\star\| + \left(\frac{1}{2}\sum_{t=1}^T (\ell_t^\star)^2\right)^{\frac{1}{2}}. \tag{24}$$

If the loss $\ell_t$ at round $t$ is greater than $1/2$, then $\tau_t \ell_t \geq \ell_t^2/(3+\gamma)$ by Eq. 15, otherwise $\tau_t = 0$. Therefore the sum of $\ell_t^2$ is less than $(3+\gamma)L_T$, which results the bound of Theorem 4. □

# D. Experiments with Binary Synthetic Sequence

## D.1. Sequence generated by single motif

We provide more comprehensive results on the synthetic binary data used in Subsection 5.1.

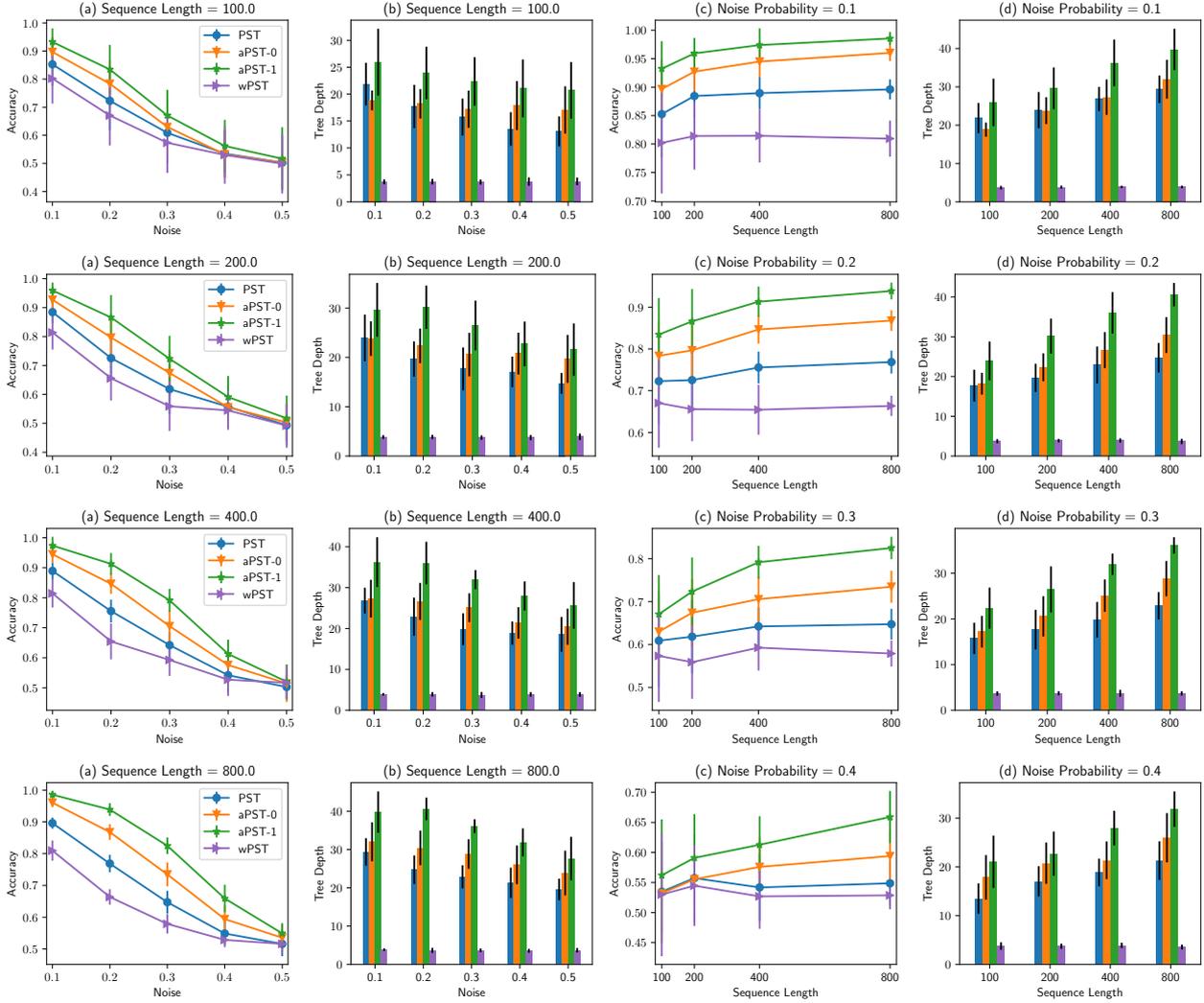

*Figure 7.* (a), (b): Prediction accuracy and tree depth of PSTs with respect to varying proportions of noise. Each row represents a different sequence length. (c), (d): Prediction accuracy and tree depth of PSTs with respect to varying lengths of sequence given the fixed noise level. Each row represents a different noise level.

### D.2. Sequence generated by mixture of motifs

In this section, we provide experiments with more complex sequence patterns than those of the main text. For the experiments, we randomly synthesize a sequence based on two motifs: $[-1, -1, +1, +1]$ and $[+1, -1, +1, -1]$. Starting from an empty sequence, on each round, we randomly choose which motif we will append at the end of the sequence, and then add a randomly corrupted motif via a fixed noise probability. Through the above process, we generate sequences of length 100, 200, 400, 800. We randomly generate 30 sequences for each length and report the average accuracy and tree depth of each model in Figure 8.

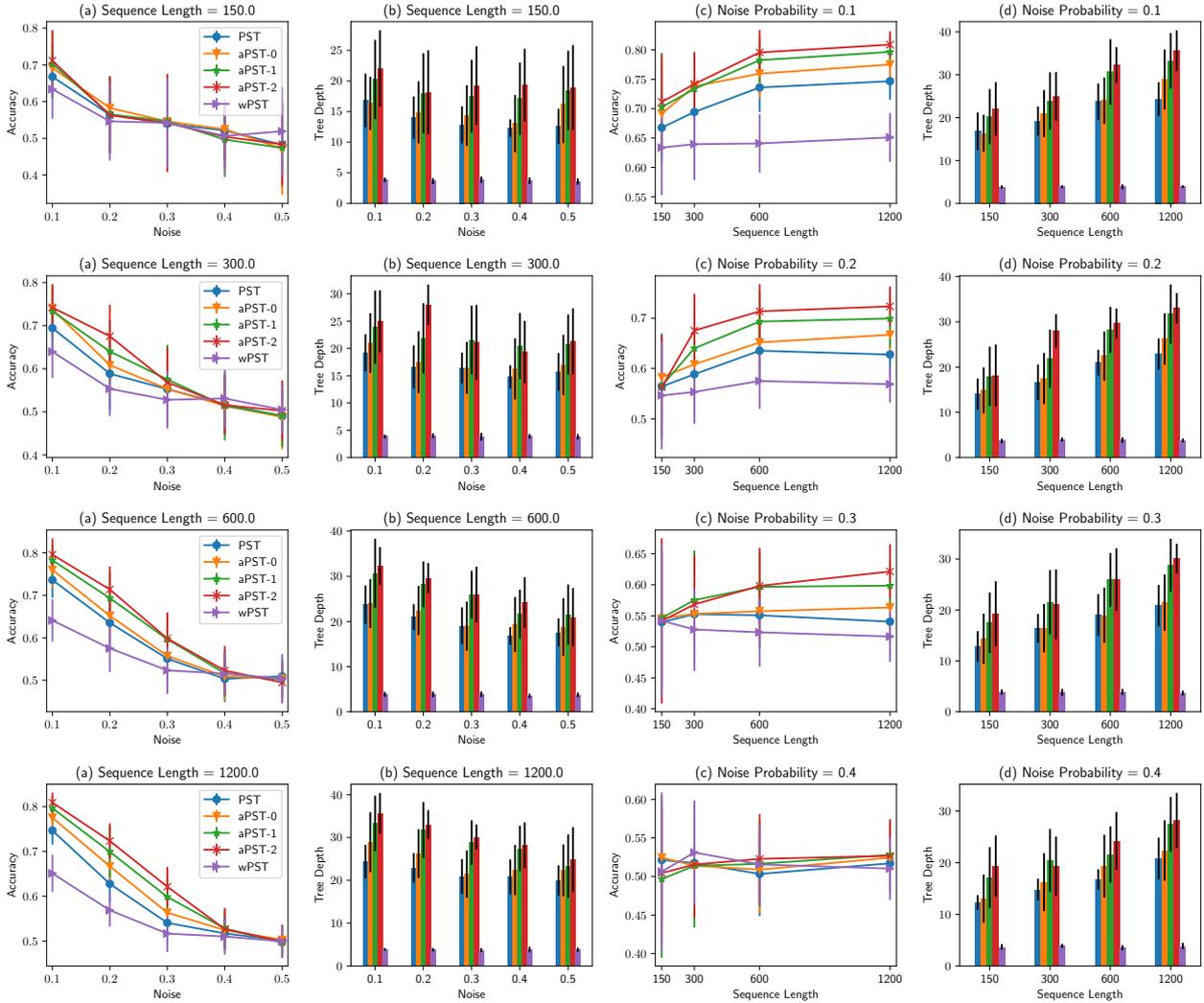

*Figure 8.* (a), (b): Prediction accuracy and tree depth of PSTs with respect to varying proportions of noise. Each row represents a different sequence length. (c), (d): Prediction accuracy and tree depth of PSTs with respect to varying lengths of sequence given the fixed noise level. Each row represents a different noise level.